\documentclass[sigconf,anonymous=false,nonacm=true]{acmart}
\usepackage{soul} 
\usepackage{cleveref} 
\Crefname{equation}{Eq.}{Eqs.}
\Crefname{figure}{Fig.}{Figs.}
\Crefname{tabular}{Tab.}{Tabs.}
\Crefname{section}{Sec.}{Secs.}

\usepackage{paralist} 
\usepackage{amsmath}
\DeclareMathOperator*{\argmin}{argmin} 
\usepackage[english]{babel}

\AtBeginDocument{%
  \providecommand\BibTeX{{%
    \normalfont B\kern-0.5em{\scshape i\kern-0.25em b}\kern-0.8em\TeX}}}

\setcopyright{acmcopyright}
\copyrightyear{2022}
\acmYear{2022}
\acmDOI{10.1145/nnnnnnn.nnnnnnn} 
\acmISBN{978-x-xxxx-xxxx-x/YY/MM} 

\acmConference[GECCO '22]{The Genetic and Evolutionary Computation Conference 2022}{July 9--13, 2022}{Boston, USA}
\acmBooktitle{GECCO '22: The Genetic and Evolutionary Computation Conference,
  July 9--13, 2022, Boston, USA}
\acmPrice{15.00}
\acmISBN{978-1-4503-XXXX-X/18/06}

\usepackage{algorithm}
\usepackage{algorithmicx}
\usepackage{algpseudocode}
\usepackage{amsmath}
 
\usepackage{babel}
\usepackage{multirow}

\begin{document}

\title{Evolvability Degeneration in Multi-Objective Genetic Programming for Symbolic Regression}

\author{Dazhuang Liu}
\email{dazhuang.liu@cwi.nl}
\author{Marco Virgolin}
\email{marco.virgolin@cwi.nl}
\affiliation{%
  \institution{Centrum Wiskunde \& Informatica}
  \city{Amsterdam}
  \country{the Netherlands}
}

\author{Tanja Alderliesten}
\email{t.alderliesten@lumc.nl}
\affiliation{%
  \institution{Leiden University Medical Center}
  \city{Leiden}
  \country{the Netherlands}
}

\author{Peter A. N. Bosman}
\email{peter.bosman@cwi.nl}
\affiliation{%
  \institution{Centrum Wiskunde \& Informatica}
  \city{Amsterdam}
  \country{the Netherlands}
}
\affiliation{%
  \institution{Delft University of Technology}
  \city{Delft}
  \country{the Netherlands}
}

\renewcommand{\shortauthors}{Liu, et al.}

\begin{abstract}
Genetic programming (GP) is one of the best approaches today to discover symbolic regression models.
To find models that trade off accuracy and complexity, the non-dominated sorting genetic algorithm II (NSGA-II) is widely used.
Unfortunately, it has been shown that NSGA-II can be inefficient: in early generations, low-complexity models over-replicate and take over most of the population.
Consequently, studies have proposed different approaches to promote diversity.
Here, we study the root of this problem, in order to design a superior approach. 
We find that the over-replication of low complexity-models is due to a lack of evolvability, i.e., the inability to produce offspring with improved accuracy.
We therefore extend NSGA-II to track, over time, the evolvability of models of different levels of complexity.
With this information, we limit how many models of each complexity level are allowed to survive the generation.
We compare this new version of NSGA-II, \emph{evoNSGA-II}, with the use of seven existing multi-objective GP approaches on ten widely-used data sets, and find that evoNSGA-II is equal or superior to using these approaches in almost all comparisons.
Furthermore, our results confirm that evoNSGA-II behaves as intended: models that are more evolvable form the majority of the population. 
\\
Code: \url{https://github.com/dzhliu/evoNSGA-II}

\end{abstract}




\maketitle

\section{Introduction}


In recent years, we are seeing a renewed interest in symbolic regression (SR), the sub-field of machine learning (ML) which concerns searching for ML models in the form of mathematical expressions~\cite{udrescu2020ai,lacava2021contemporary,zhang2021rl,dascoli2022deep}.
These models are appealing because, by their very nature, they stand a chance of being interpretable. 
This is increasingly considered important, e.g., to ensure that ML is used in a fair and responsible manner~\cite{adadi2018peeking,2019DARPA,la2020genetic}.

Today, genetic programming (GP)~\cite{koza1992genetic} is one of the best approaches to discover SR models~\cite{lacava2021contemporary}.
GP is a bio-inspired meta-heuristic that works by \emph{evolving} a population of solutions that, differently from traditional genetic algorithms, need be \emph{executed} to be evaluated, i.e., they are programs. 
In the case of SR, the solutions evolved by GP encode functions as symbolic models that are evaluated in terms of their accuracy in fitting a (training) data set~\cite{koza1992genetic}.
However, when maximizing accuracy alone, GP tends to generate solutions that become unnecessarily large in the number of components (arithmetic operations, variables, constants, etc.), a phenomenon known as \emph{bloat}, which harms interpretability~\cite{bloatcontrol}.

To deal with this problem, GP can be set to optimize different objectives at the same time. 
Multi-objective GP (MOGP) is  typically used with the intention to search for solutions with different trade-offs between accuracy and interpretability ~\cite{kommenda2016evolving}. 
At the end of a single run of MOGP, decision makers can choose the model that strikes the right balance between accuracy and interpretability.
Since interpretability is hard or impossible to define (in general terms)~\cite{lipton2018mythos,virgolin2021model}, the common way by which interpretability in pursued in MOGP for SR is by minimization of solution size (or derivations thereof, see e.g., the related work section in~\cite{Virgolin2020PPSN}), i.e., the number of components that constitutes the solution.
Minimizing size is typically in conflict with maximizing accuracy in (MO)GP, because (MO)GP typically discovers better solutions by refining the function approximation they represent, i.e., by incorporating additional components~\cite{langdon2021genetic}.

Just like solution size is the objective that is typically adopted, the second version of the non-dominated sorting genetic algorithm~\cite{nsgaii} (NSGA-II) is the most adopted framework to realize MOGP.
Unfortunately, as it has been shown by several works before~\cite{DeJong2003,populationcollapse2007,alphadominance,adaptivealphadominance} and is confirmed once more in this paper, NSGA-II can be inefficient when adopted for MOGP when one of the objectives is solution size.
In particular, small solutions are observed to take over the majority of the population in a few generations, while larger and more accurate solutions are hardly discovered.

In this paper, we tackle this problem at its root. 
Specifically, we identify that the reason why small solutions over-replicate and hamper the discovery of larger but more accurate solutions is the fact that, besides obviously minimizing size well and thus having high chances of survival, small solutions lack \emph{evolvability}.
Here, by evolvability of a solution we mean the likelihood that variation (e.g., subtree crossover and mutation) produces a relatively accurate offspring when using that solution as a parent.
We call this cause of inefficiency of NSGA-II \emph{evolvability degeneration}.
Consequently, we present a new algorithm, named \emph{evoNSGA-II}, which improves upon standard NSGA-II by restraining the over-replication of solutions whose size is identified to be unhelpful in terms of discovering more accurate solutions.
    Thanks to this, we find evoNSGA-II to be far more efficient than NSGA-II as well as other algorithms designed to deal with this issue.


\section{Background \& related work}\label{background}

\subsection{Brief recall on SR and (MO)GP}

In SR, we seek a model (or equivalently, function approximation) $f$ that is accurate in terms of fitting a given data set.
Accuracy is typically measured in terms of minimizing a loss function, such as the mean-squared-error (MSE).
Formally, given a data set $\mathcal{D} = \{ (\mathbf{x}_i, y_i ) \}^n_{i=1}$, where $n$ is the number of observations, $\mathbf{x}_i \in \mathbb{R}^d$ is the vector of \emph{d} feature values $\mathbf{x}_i = \left( x^{(1)}_i, \dots, x^{(d)}_i \right)^\top$, and $y_{i} \in \mathbb{R}$ the label or target variable, we seek an optimal $f^\star$ such that:
\begin{equation*}
\label{eq:loss}
     f^\star := \argmin_{f\in F} \left\{\text{MSE} \left(\mathcal{D}, f\right) \right\}= \argmin_{f \in F}\left\{ \frac{\sum_{i=1}^n \left( y_i - f(\mathbf{x}_i) \right)^2}{n} \right\}.
\end{equation*}
An SR algorithm searches in the space of functions $F$ that is defined in terms of an \emph{encoding} (see next paragraph), and what atomic sub-functions ($+$, $-$, $\times$, $\div$, $\exp$, $\log$, etc.), variables ($x^{(1)}$, $x^{(2)}$, etc.), and constants ($\frac{1}{2}$, $-\pi$, $42$, etc.) appear in what order in that encoding. 
Alongside maximizing accuracy, we wish the model to be interpretable. 
Various metrics have been proposed to seek interpretable/simpler models, see e.g., ~\cite{Virgolin2020PPSN, Vladislavleva2009}. 
However, reducing  model size remains a simple and popular approach (e.g., it was recently used in a large SR benchmark~\cite{lacava2021contemporary}).

GP is a popular and often top-performing method for SR~\cite{lacava2021contemporary}. 
In this work, we adopt traditional GP, where solutions are encoded by trees in which each node contains one of the possible sub-functions, variables, and constants~\cite{koza1992genetic,poli2008field}.
To discover of multiple solutions with trade-offs between accuracy and interpretability,
GP is set to work in a multi-objective fashion (MOGP), where the concept of Pareto-dominance is used to rank solutions.
Specifically, we say that solution $A$ Pareto-dominates solution $B$ if $A$ is \emph{equal or better} than $B$ in all objectives, and strictly better in at least one objective. 
The outcome of MOGP is the best-found \emph{front}, i.e., the set of solutions that are not Pareto-dominated by any other ever found.

NSGA-II is widely considered to be the most popular multi-objective evolutionary algorithm (MOEA). 
We conducted a small literature survey to assess whether this is indeed the case for MOGP. 
We detail how the survey was conducted in the Appendix. 
We found that, in the last five years, NSGA-II was typically adopted as MOGP algorithm in approximately $70\%$ of the works that we surveyed, either as the main algorithm or as a baseline.
We thus believe that our intent of improving NSGA-II for MOGP is amply justified.


\subsection{Prior works on improving NSGA-II for GP}
\label{sec:prior-attempts}

Several works in the literature have identified the problem of small solutions over-replicating and hampering further evolution, which we refer to as \emph{evolvability degeneration}.
A very-closely related concept was discovered almost twenty years ago in~\cite{DeJong2003}, and termed later as \emph{population collapse}~\cite{populationcollapse2007}. 
Population collapse refers to the process where the entire population converges to copies of a single solution that has a single component, i.e., the population is unable to evolve any further.
As it will be shown in this paper (in \Cref{section:evolvDegenWhyHappened}), the behavior we observe is less extreme: 
even though copies of small solutions do initially occupy most of the population in early generations, NSGA-II remains able to recover, i.e., larger solutions are discovered later on, albeit at a very slow rate. 
To prevent population collapse, the use of a diversity preservation mechanism is advised in~\cite{DeJong2003}.
Instead, in~\cite{populationcollapse2007} it is argued that employing mutation is enough.
Here, we find that even if one employs mutation, NSGA-II still suffers from evolvability degeneration.

Other works have also noted, and proposed means to deal with, the problem of small solutions flooding the population.
In~\cite{bleuler2001multiobjective}, it is proposed to use SPEA2 for MOGP\cite{zitzler2001spea2}, to overcome the problem just mentioned as well as bloat.
SPEA2, which we also consider in our experiments, works in a fundamentally different way than NSGA-II.
For example, SPEA2 maintains two separate populations during the search, and measures the performance of a solution based on how many solutions are dominated by that solution.


Recently,~\cite{alphadominance} and~\cite{adaptivealphadominance} explored the idea of using $\alpha$-dominance.
Instead of the original objectives (here, accuracy and size), these algorithms use linear combinations of the original objectives which are weighted by coefficients ($\alpha$) that vary over time, so as to be able to put more pressure on finding solutions of a certain trade-off.
In particular, $\alpha$ is adapted to increase the importance of accuracy over the importance of size.
In the first work~\cite{alphadominance}, fixed schedules are considered to adapt $\alpha$, according to a function of the number of generations that is linear, a cosine, or a sigmoid. 
In the second work~\cite{adaptivealphadominance}, $\alpha$ is adapted dynamically based on the state of the population: if more small than accurate (and vice versa, accurate than small) solutions are detected, then $\alpha$ is adapted to give more weight to accuracy (respectively, to size).

For NSGA-II applied to discrete optimization, in~\cite{HisaoOverlapping2005} strategies are explored to remove duplicate solutions from the population.
One such strategy is used for MOGP in~\cite{virgolin2021model}, where NSGA-II is modified so that duplicate solutions are assigned the lowest priority to survive selection. 
Together with classic NSGA-II, SPEA2, and the $\alpha$-dominance based algorithms, we also include this algorithm in our comparisons.

To the best of our knowledge, our work differs from the previous ones because it makes an explicit link between the over-replication of small solutions and their lack of evolvability, and proposes an algorithm that uses this information to improve the search.

\section{Evolvability degeneration}
\label{section:evolvDegenWhyHappened}
In this section, we analyze the phenomenon of evolvability degeneration in NSGA-II for MOGP.
First, we describe it by considering a use case.
Then, we show what causes it.
The latter is done by means of an experiment in which we trace how solutions of different sizes contribute to finding offspring solutions that are relatively accurate.

\subsection{Over-replication of small solutions}
\label{sec:over-replication}
We begin by reporting how the size of solutions changes over time when using NSGA-II on an examplary use case.
The parameter settings for NSGA-II are those in bold font in \Cref{parametersetting}, except for the population size, which is set to 500.
We show the behavior of NSGA-II on the data set Airfoil (see Sec.~B of the Appendix).
We use this data set as a recurring example for no particular reason other than it being first in alphabetic order among the data sets we considered; we observe similar trends also on the other data sets.

\begin{figure}[]
\centering
\includegraphics[width=0.7\linewidth]{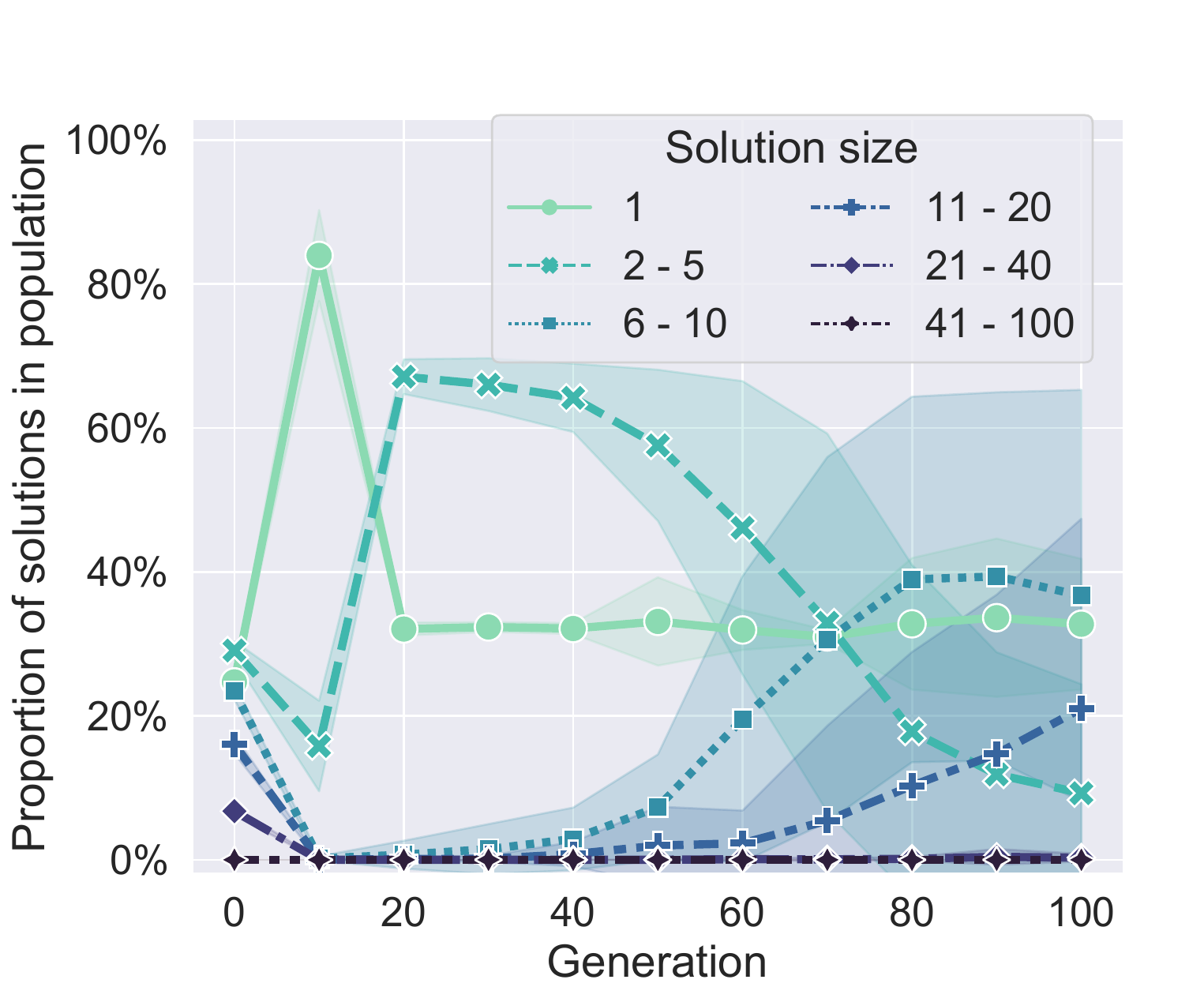}
\vspace{-2mm}
\caption{Proportion of solutions of different sizes during the evolution for 30 runs of NSGA-II on Airfoil. 
Lines indicate means and shaded areas represent standard deviations. 
Note the exponential scaling of solution size intervals.}
\label{fig:NSGAII-forPopulationCollapseAnalysis}
\end{figure}

\Cref{fig:NSGAII-forPopulationCollapseAnalysis} shows that, at the initial stages of the evolution, the proportion of small solutions grows to occupy the majority of the population.
Only later small solutions start to diminish and slightly larger solutions start to appear and compete. 
However, the largest solutions, in this case the ones with more than $20$ nodes, are basically not discovered.
Importantly, the solutions of size one always occupy a rather large portion of the population (above $30\%$). 
This abundance of small solutions can be explained by the fact that, reasonably, small solutions of relatively high accuracy and duplicates thereof are produced by GP relatively quickly; in particular, before larger and more accurate solutions are discovered.
Because of how NSGA-II works, solutions that have the best-so-far accuracy for any given size are set to survive the generation with high priority, no matter if they are duplicates or not.
Now, this abundance of small solutions would not necessarily be a problem if small solutions would represent fertile grounds to discover larger, more accurate solutions.
In the next section, we show that this is not the case.

\subsection{Evolvability of small and large solutions}
\label{sec:evolv-small-solutions}

A simple way to understand whether evolution stagnates or proceeds well is to measure evolvability in terms of the frequency by which well-performing offspring solutions are discovered.
Here, we particularly want to measure the frequency with which solutions of different size contribute to offspring solutions with an accuracy that is relatively high.
Since we aim to improve NSGA-II, we would ideally do this within an NSGA-II evolution.
However, as shown in \Cref{fig:NSGAII-forPopulationCollapseAnalysis}, larger solutions are hardly ever discovered, making it impossible for us to estimate their evolvability.
Thus, we design a workflow to collect enough solutions of various sizes. 

First, we repeatedly run single-objective GP, 100 times, with different maximal size limitations, for up to a certain number of generations (e.g., 40). 
This allows us to collect best-found solutions of various sizes which are relatively accurate, and can be imagined to contribute to a best-found front at a certain stage of an ``ideal'' NSGA-II evolution, where evolvability degeneration does not occur.
Second, we collect these solutions in different buckets, based on their size. 
We also record the $90^\textit{th}$ percentile of accuracy ($\emph{acc}_{90}$) out of all solutions collected, irrespective of their size; we use this information later.
Next, for each bucket, we repeatedly (100 times) take a random solution to act as parent, and generate an offspring solution via subtree mutation.
We do the same for subtree crossover, this time considering pair of buckets and, importantly, generating a \emph{single} offspring instead of two (this is rather common in GP~\cite{poli2008field}).
Specifically, the offspring is generated by cloning the first parent and transplanting a random subtree from the second parent (which from now on will be called \emph{donor} to avoid confusion) to replace a random subtree of the first parent (which from now on will be simply called \emph{parent}).
We perform crossover this way because, for a sufficiently large parent, in expectation the majority of the nodes in the offspring comes from the parent instead of the donor; this may play an important role in terms of evolvability.
Lastly, we measure how frequently parents of different sizes produce relatively accurate offspring, using $\emph{acc}_{90}$ as a threshold.

\begin{figure}[]
\centering
\includegraphics[width=0.9\linewidth]{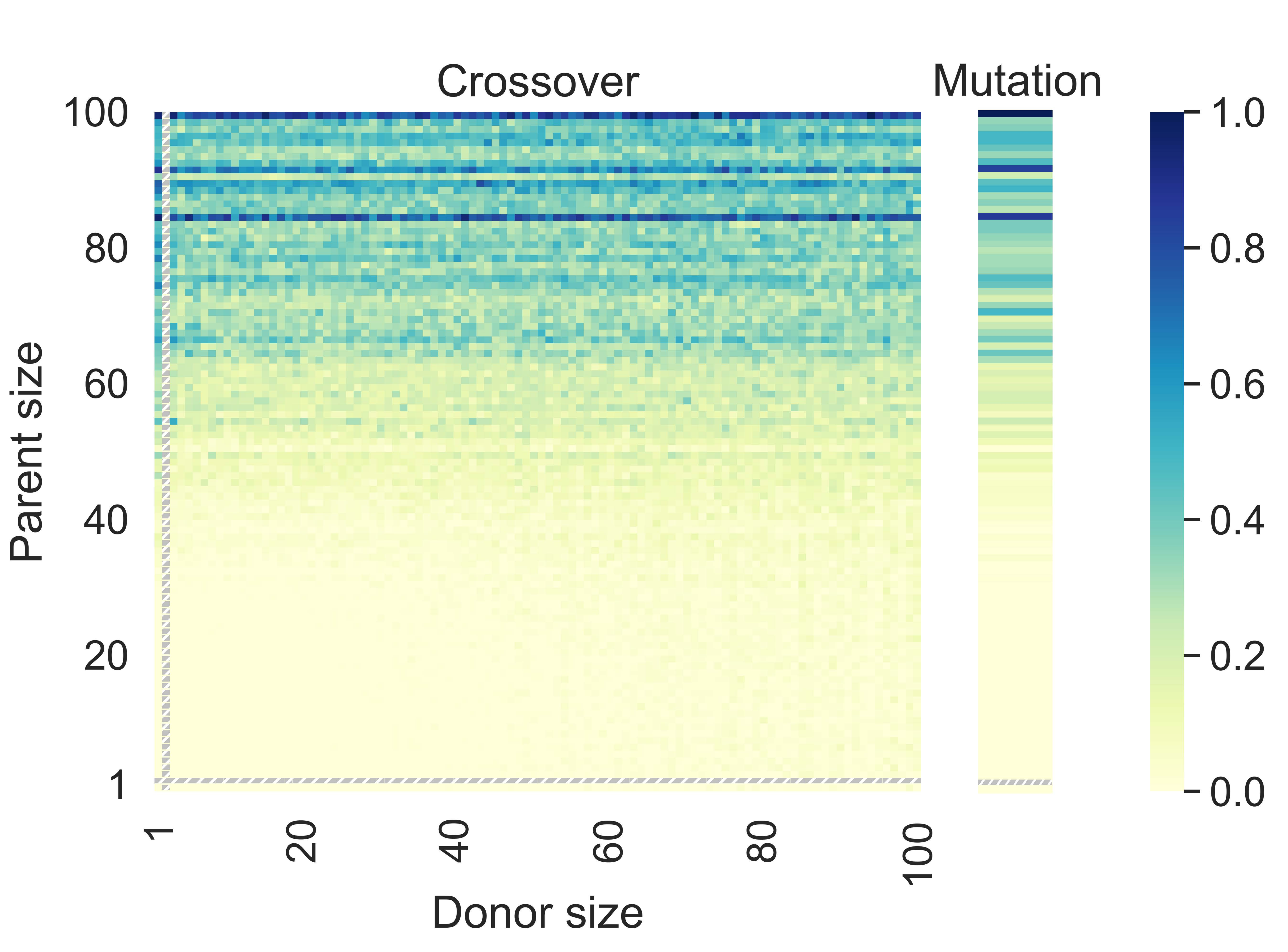}
\caption{Frequency (normalized between min and max, color coded as depicted by the legend on the right) of producing an offspring with a good accuracy (above the $90^\textit{th}$ percentile of those obtained in all runs) based on the size of the parent and donor solutions. 
Left: For subtree crossover. Right: For subtree mutation. 
Note that a solution of size 2 (to be used as a parent or donor) was never returned by single-objective GP because a more accurate solution of size 1 exists and was systematically discovered.}
\label{heatmap-crossovermutation}
\end{figure}

We apply the proposed workflow and display the result in  \Cref{heatmap-crossovermutation} (all the parameter settings are as per \Cref{sec:over-replication}), which concerns Airfoil and best-found solutions at generation 40.
We remark that we repeated the same approach on the second data set we consider, Boston, as well as with other termination limits (generation 10, 20, and 30), and means of assessing whether an offspring is relatively accurate (e.g., with respect to the accuracy of the parent); 
we observed the same general trends as shown in \Cref{heatmap-crossovermutation}. 
Note that the heat-map for subtree crossover is not symmetric due to the reasons explained in the previous paragraph.
The frequencies found for subtree crossover indicate that the parent needs to be sufficiently large for variation to be successful with large probability, while the donor can be of any size. 
Similarly, also for mutation larger solutions are more evolvable. 

The result just shown confirms our hypothesis that smaller solutions hamper the search. 
Therefore, the fact that in the early stages of an NSGA-II run, the population is flooded by copies of small solutions, is highly undesirable. 
We remark that penalizing duplicates altogether, as in fact was done in some earlier approaches (see \Cref{sec:prior-attempts}), is not necessarily the optimal strategy.
In fact, having duplicates of highly-evolvable solutions may be the best option.
This idea is explored in our algorithm, presented in the next section.



\section{Improving NSGA-II based on evolvability}\label{Evolvability-based NSGA-II}

We now present our proposal to improve NSGA-II, i.e., evoNSGA-II.
Since evoNSGA-II mostly follows NSGA-II, we begin by recalling the workings of NSGA-II.
Next, we explain what is new in evoNSGA-II, i.e., the estimation of the evolvability of solutions of different size, and the use of this information to prevent the over-replication of solutions with low evolvability.

\subsection{NSGA-II}
\label{sec:workings-nsga-ii}

\Cref{NSGAIIEVONSGAII} shows the pseudo-code of NSGA-II, as well as that of evoNSGA-II: In fact, the only change we apply is regarded to how the population is updated at the end of a generation.
In every generation of (evo)NSGA-II, firstly an offspring population $\mathcal{O}$ is generated from promising solutions of the current population $\mathcal{P}$. 
Promising solutions are typically chosen with tournament selection, and then undergo variation, typically by means of subtree crossover and subtree mutation. 
In (evo)NSGA-II, tournament selection compares solutions based on their non-domination \emph{rank} (explained below) and, if the solutions share the same rank, based on their \emph{crowding distance} (explained below too).

Next, $\mathcal{P}$ and $\mathcal{O}$ are merged and undergo non-dominated sorting.
Non-dominated sorting is a process that subdivides all solutions into layers called \emph{fronts}, such that for any two solutions in a same front, those two solutions do not Pareto-dominate each other; 
moreover, for each solution in the $i^\textit{th}$ front, there exists at least one solution in the $(i-1)^\textit{th}$ front that Pareto-dominates it.
The rank of a solution represents the front to which that solution belongs, rank $1$ being the best.
The algorithm proceeds by parsing each front and assigning to each solution in that front a \emph{crowding distance}.
The crowding distance is a measure of sparseness (the more a solution is isolated the better) that is computed in the objective space using the L1 norm.
A solution for which an objective has the maximum value for that front is assigned an infinite (and thus best) crowding distance.

Finally, the population is updated for the next generation, using an NSGA-II-specific form of \emph{truncation} selection.
This is where NSGA-II and evoNSGA-II differ.
In NSGA-II, the new population is formed by selecting the solutions with rank 1, then those with rank 2, and so on, until the selection of all solutions with a certain rank would result in exceeding the population size.
In that case, the crowding distance is used to discern which subset of solutions with that certain rank still to select for the new population.
The remaining solutions are discarded.

\begin{algorithm}
    \caption{Workflow of NSGA-II and evoNSGA-II\\ \small \textbf{Note}:
    \emph{Truncation} is the only step that is different between the two.}
    \begin{algorithmic}[1]
        \Require \emph{Pop\_size, stop\_criteria}
        \State $\mathcal{P} \leftarrow$ \Call{Initialize\_population}{Pop\_size}
        \State \Call{Evaluate}{$\mathcal{P}$}
        \State \emph{Fronts}$\leftarrow$\Call{Fast\_non-dominated\_sorting}{$\mathcal{P}$}
            \For{\emph{front} in \emph{Fronts}}
                \State \Call{Crowding\_distance}{\emph{front}}
            \EndFor
        \While{$\neg$ \emph{stop\_criteria}}
            \State $\mathcal{P}^\prime \leftarrow$ \Call{Tournament}{$\mathcal{P}$}
            \State $\mathcal{O}$ $\leftarrow$  \Call{Variation}{$\mathcal{P}^{'}$}
            \State \Call{Evaluate}{$\mathcal{O}$}
            \State \emph{Fronts} $\leftarrow$  \Call{Fast\_non-dominated\_sorting}{$\mathcal{P} \cup \mathcal{O}$}
            \For{\emph{front} in \emph{Fronts}}
                \State \Call{Crowding\_distance}{\emph{front}}
            \EndFor
            \State \emph{$\mathcal{P}$} $\leftarrow$  \boxed{\Call{Truncation}{\emph{Fronts}}}
        \EndWhile
    \end{algorithmic}
    \label{NSGAIIEVONSGAII}
\end{algorithm}

evoNSGA-II additionally uses estimates of evolvability for each size of solution to decide whether a solution should be selected.
Specifically, we generate a table of bounds $\mathcal{B}$ that tells how many solutions of a certain size can be selected in the truncation selection step. 
This way we can prevent the over-replication of small, non-evolvable solutions.
We proceed by explaining how $\mathcal{B}$ is built.

\subsection{Construction of $\mathcal{B}$}
We keep track of the evolvability of solutions in terms of their capability of generating accurate offspring of different sizes.
Namely, we build a table $\mathcal{B}$ containing pairs ($s$,$b$), where $s$ is a size and $b$ is a bound on the number of times that solutions of size $s$ can be selected by truncation selection to form the new population of evoNSGA-II.
We want the number $b$ to be proportionate to the (estimated) evolvability of the solutions of size $s$.

\Cref{alg:bounds} shows the construction of $\mathcal{B}$ in detail.
For each offspring, the size of \emph{its parent} $s$ is considered. 
Then, a counter (\emph{successes}) that is dedicated to that $s$ is increased if the accuracy of the offspring is larger than that of the median accuracy computed over $\mathcal{P}$ (we choose the median over the mean because outliers are common in GP for SR). 
Note that we do not need to re-compute the accuracy of solutions, as they can simply be cached when solutions are evaluated.
We also keep track of the number of offsprings that was generated from parents of size $s$ (\emph{attempts}).
Finally, a simple measure of evolvability is computed for $s$, as the ratio between the number of successes and the number of attempts.
This ratio is in $[0,1]$ and the larger its value, the better it is.
We fill $\mathcal{B}$ with these ratios, for each size.

Recall that we wish to use $\mathcal{B}$ in the truncation selection process, which is applied to $\mathcal{P} \cup \mathcal{O}$.
Importantly, $\mathcal{P} \cup \mathcal{O}$ contains both solutions that were not selected as parents, and offspring solutions: 
for those, there may exist a size that is not in $\mathcal{B}$, i.e., for which we have no information on its evolvability.
Therefore, we artificially fill this information for potentially-missing sizes in $\mathcal{B}$ (line 13).
Namely, for each missing size, we take the weighted average of the ratios observed for the closest smaller and closest larger size.
Last but not least, we perform a normalization step on $\mathcal{B}$, transforming the ratios so that their sum amounts to the population size ($|\mathcal{P}|$).
This way, for any size $s$, $\mathcal{B}[s]$ defines how many solutions should be selected at most.
In the next section, we illustrate how $\mathcal{B}$ is used in the truncation selection of evoNSGA-II.

\begin{algorithm}
    \caption{Build\_$\mathcal{B}$}
    \begin{algorithmic}[1]
       \Require $\mathcal{P}$, $\mathcal{O}$
       \State \emph{max\_size} $\gets$ \Call{Max\_size}{$\mathcal{P} \cup \mathcal{O}$}
       \State \emph{attempts}[i] $\gets 0 \text{ for } i \in \left\{1, \dots, \textit{max\_size} \right\}$
       \State \emph{successes}[i] $\gets 0 \text{ for } i \in \left\{1, \dots, \textit{max\_size}\right\}$
       \State \emph{median\_accuracy} $\leftarrow$ \Call{Median\_accuracy}{$\mathcal{P}$}
       \For{$o \in \mathcal{O}$}
           \State $s \gets \Call{Fetch\_parent\_size}{o}$
           \If{$ \Call{Accuracy}{o} > \emph{median\_accuracy}$}
               \State $\emph{successes}[s] \gets \emph{successes}[s] + 1$
           \EndIf
           \State $\emph{attempts}[s] \gets \emph{attempts}[s] + 1$
       \EndFor
       \State $\mathcal{B}[i] \leftarrow \frac{\textit{successes}[i]}{\textit{attempts}[i]} \text{ for } i \in \{1, \dots, \emph{max\_size} \} :  \emph{attempts}[\emph{i}] \neq 0 $
        \State $\mathcal{B} \leftarrow  \Call{Fill\_missing\_sizes}{\mathcal{B}, \mathcal{P}, \mathcal{O}}$ 
        \State $\mathcal{B} \leftarrow  \Call{Normalize}{\mathcal{B}, |\mathcal{P}|}$ 
        \State $\Return{~\mathcal{B}}$
    \end{algorithmic}
    \label{alg:bounds}
\end{algorithm}




\subsection{Use of $\mathcal{B}$ during truncation selection}

The way truncation selection works in evoNSGA-II is the same as in NSGA-II (as described before, in \Cref{sec:workings-nsga-ii}), except for the fact that we will now use $\mathcal{B}$ to decide how many solutions of a certain size can be selected.
We build $\mathcal{B}$ after the offspring population $\mathcal{O}$ has been evaluated, so that it is ready to be used for truncation selection.
Like in NSGA-II, our truncation selection parses the solutions progressively, based on their rank.
Different from NSGA-II, we do not immediately copy the solution that is currently in consideration; 
first, we consider the size of $s$ of that solution, and the respective bound $\mathcal{B}[s]$.
If the number of solutions of size $s$ copied so far is less or equal to $\mathcal{B}[s]$, then the solution is selected; 
else, the solution is skipped, and the next solution is considered.

It can happen that, in \Cref{alg:bounds}, large evolvability values are estimated for sizes for which there is a limited number of solutions in $\mathcal{P} \cup \mathcal{O}$, while low values are estimated for sizes for which there is an abundant number of solutions.
Consequently, to respect the bounds in $\mathcal{B}$, the number of selected solutions may be lower than the population size.
If that happens, we reset the counters for how many solutions of each size have been copied, and start the truncation selection process anew, from rank 1 onwards.
This way, even if the bound for the size $s$ is exceeded, we maintain an approximate proportionality between estimated evolvability of $s$ and the number of selected solutions of size $s$.

Lastly, we remark that a single generation of evoNSGA-II is basically as fast as NSGA-II, as it entails minimial overhead.
In fact, from the perspective of computational complexity, all operations needed to build and use $\mathcal{B}$ are linear in the population size, and thus subsumed by the complexity of other operations, particularly non-dominated sorting and evaluation of accuracy.



\section{Experimental setup}\label{section:experimental-setup}
We consider ten data sets that are commonly used in recent literature on GP for SR. 
The information for these data sets is reported in Appendix B due to space limitations.
For any run, we use a traditional Monte-Carlo split of the data set into training and test set, with respective proportions of 75-25\%.
Moreover, all data sets are standardized (based on the information in the training set) by subtracting the mean and dividing by the standard deviation for each feature separately, as advised in~\cite{dick2020feature}.

For comparison, we consider seven algorithms besides evoNSGA-II: classic NSGA-II~\cite{nsgaii}, SPEA2~\cite{zitzler2001spea2}, $\alpha$-dominance-based NSGA-II~\cite{alphadominance} with $\alpha$ varied with a linear ($\alpha$-dom. lin.), cosine~($\alpha$-dom. cos.), or sigmoid~($\alpha$-dom. sig.) schedule, as well as its adaptive version~\cite{adaptivealphadominance} (Adap.~$\alpha$-dom.), and a simple extension of NSGA-II as mentioned in~\cite{virgolin2021model}, where non-dominated sorting assigns an artificial worst-possible rank to duplicate solutions.
We refer to the latter as NSGA-II with penalization of duplicates, NSGA-II+PD in short.
For each algorithm, we keep track of the best-ever found non-dominated solutions (with respect to the training set) in an external archive, and return that archive at the end of the evolution.

Solutions are evaluated in terms of accuracy (to maximize) and size (to minimize). 
To maximize accuracy, we minimize the MSE (\Cref{eq:loss}) augmented by linear scaling~\cite{keijzer2003improving}.
Linear scaling effectively enables to optimize in terms of a form of absolute correlation to the target variable $y$, typically causing a large improvement when GP is applied on real-world SR data sets~\cite{virgolin2019linear}. 
Many state-of-the-art GP algorithms use linear scaling during the evolution~\cite{lacava2021contemporary}.

To evaluate the quality of multi-objective search, we compute the hypervolume (HV) of the archive of best-found non-dominated solutions~\cite{Hypervolume}. 
The HV indicates, for a set of solutions, the area in objective that is Pareto-dominated by that set of solutions, bounded by a reference point.
The reference point represents an artificial solution with (very) poor performance in terms of all considered objectives, and should be chosen to be commensurate to the ranges of the objectives at play.
We set the reference point to be $(1.1,1.1)$ (meaning that the best-possible HV will be $1.1^2=1.21$) and normalize the MSE and size to be within 0 and 1.
Even though the MSE would normally be unbounded from above, performing linear scaling guarantees that the maximal training error corresponds to predicting the mean of $y$;
thus we can achieve the desired normalization by dividing by the variance of $y$.
Regarding size, since very large solutions will likely not be interpretable, we enforce a maximal solution size of $100$ (see \Cref{parametersetting}) by deleting any offspring that exceeds that limit (and cloning the parent in its place); size is then normalized by dividing by $100$.

We perform $30$ repetitions for each run, to account for the randomness of train-test splitting and the stochasticity inherent to GP.
We strive to present our results in terms of a typical parameter configuration that appears often in GP literature.
To that end, we actually consider a number of typical configurations; see \Cref{parametersetting}, where some parameters have different possible settings (namely, population size, tournament size, and proportion between crossover and mutation).
Note that starred operators (e.g., $\div^*$) implement protection and ephemeral random constants (ERC)~\cite{poli2008field} are sampled within $\mathcal{U}(-5,+5) \times \max_{i,j} |x^{(j)}_i|$.
For each algorithm, we find the configuration that leads to the \emph{average} performance for that algorithm (on the training set).
The configuration that leads to the \emph{average} performance for an algorithm is found as follows.
First, for each data set, we consider the training HV (averaged across 30 runs) obtained on that data set by the different configurations.
Configurations are sorted based on their HV, and their sort order is taken as a score.
Next, an overall, single score is assigned to each configuration, by averaging the scores across the data sets of that configuration.
Finally, we select the configuration whose overall score is closest to the one obtained by averaging the scores of all configurations.
For example, the parameter settings in bold in \Cref{parametersetting} represent the configuration obtained for evoNSGA-II; The configurations for the other algorithms are reported in Appendix C.

\begin{table}
\caption{Parameter settings considered for \emph{evoNSGA-II} and the other algorithms.
Tournament size of 1 corresponds to random parent selection.
SPEA2 does not employ tournament selection.
For parameters with multiple possible settings (i.e., the first three), the settings in bold correspond to those that result in evoNSGA-II achieving the average overall performance in terms of hyper-volume on the training set.
}
\label{parametersetting}
\begin{tabular}{lc}
\toprule
Parameter & Considered settings\\
\midrule
Population size & 250, 500, \textbf{1000}, 2000, 5000\\
Tournament size & 1, \textbf{2}, 7\\
Crossover-mutation proportion & 0.5-0.5, \textbf{0.9-0.1}\\
\hline
Initialization & Ramped half-\&-half (2--6)\\
Maximum solution size & 100\\
Function set & \small $\{+,-,\times,\div^*,\sqrt{}^*,\log^* \}$\\
Terminal set & \small $\{x^{(1)}, \dots, x^{(d)}, \text{ERC} \}$\\
\bottomrule
\end{tabular}
\end{table}

We use the Mann-Whitney-U test~\cite{mann1947test} to assess whether the distribution of HVs obtained by an algorithm is better than that of another, determining significance for $p\text{-value} < 0.05$, with Bonferroni correction~\cite{bonferroni1936teoria}.

\begin{table*}
\caption{Mean (standard deviation) of the HV computed on the training set for 30 runs of the considered algorithms.
This table corresponds to the settings in bold in \Cref{parametersetting}.
The symbols $+,-,=$ indicate, for each algorithm other than evoNSGA-II, whether the corresponding distribution of results for evoNSGA-II is, respectively, significantly better, worse, or not significantly different.
The last row summarizes this information.
}
\resizebox{\textwidth}{!}{
\begin{tabular}{lcccccccc}
\toprule
  Data set & evoNSGA-II & Adap. $\alpha$-dom. & $\alpha$-dom. cos. & $\alpha$-dom. lin. & $\alpha$-dom. sig. & NSGA-II & NSGA-II+PD & SPEA2 \\
\midrule
Airfoil       & 0.799(0.021) & 0.595(0.023)- & 0.744(0.016)- & 0.745(0.019)- & 0.736(0.016)- & 0.652(0.035)- & 0.780(0.017)- & 0.624(0.049)- \\
Boston        & 1.023(0.012) & 0.895(0.024)- & 0.984(0.010)- & 0.986(0.014)- & 0.980(0.014)- & 0.954(0.015)- & 1.017(0.010)= & 0.929(0.019)- \\
Concrete      & 0.939(0.018) & 0.684(0.033)- & 0.884(0.029)- & 0.876(0.025)- & 0.864(0.030)- & 0.792(0.037)- & 0.941(0.018)= & 0.725(0.053)- \\
Dow chemical   & 0.972(0.013) & 0.714(0.043)- & 0.914(0.013)- & 0.914(0.018)- & 0.908(0.023)- & 0.779(0.057)- & 0.977(0.007)= & 0.836(0.037)- \\
Energy: cooling & 1.119(0.008) & 1.043(0.013)- & 1.094(0.010)- & 1.094(0.010)- & 1.088(0.014)- & 1.058(0.009)- & 1.097(0.015)- & 1.040(0.013)- \\
Energy: heating & 1.152(0.004) & 1.076(0.017)- & 1.125(0.010)- & 1.125(0.008)- & 1.117(0.012)- & 1.098(0.010)- & 1.131(0.010)- & 1.054(0.014)- \\
Tower         & 1.027(0.013) & 0.824(0.046)- & 0.994(0.012)- & 0.985(0.027)- & 0.984(0.024)- & 0.941(0.030)- & 1.029(0.006)= & 0.867(0.047)- \\
Wine: red       & 0.513(0.005) & 0.446(0.007)- & 0.491(0.004)- & 0.490(0.008)- & 0.486(0.008)- & 0.461(0.009)- & 0.509(0.004)- & 0.459(0.009)- \\
Wine: white     & 0.449(0.005) & 0.378(0.008)- & 0.426(0.007)- & 0.422(0.008)- & 0.416(0.009)- & 0.403(0.007)- & 0.446(0.004)= & 0.387(0.010)- \\
Yacht         & 1.177(0.004) & 1.141(0.019)- & 1.174(0.001)- & 1.174(0.001)- & 1.174(0.001)- & 1.163(0.013)- & 1.178(0.001)+ & 1.150(0.011)- \\
\hline
Total $+/-/=$ & --- &  0/10/0 & 0/10/0 & 0/10/0 & 0/10/0 & 0/10/0 & 1/4/5 & 0/10/0 \\
\bottomrule
\end{tabular}
}
\label{resulttraining}
\end{table*}

\begin{table*}
\caption{Results for the test set, formatting similar to that of \Cref{resulttraining}.}
\resizebox{\textwidth}{!}{
\begin{tabular}{lcccccccc}
\toprule
 Data set & evoNSGA-II & Adap. $\alpha$-dom. & $\alpha$-dom. cos. & $\alpha$-dom. lin. & $\alpha$-dom. sig. & NSGA-II & NSGA-II+PD & SPEA2 \\
\midrule 
Airfoil       & 0.813(0.021) & 0.669(0.027)- & 0.781(0.018)- & 0.782(0.017)- & 0.781(0.013)- & 0.708(0.030)- & 0.795(0.019)- & 0.690(0.041)- \\
Boston        & 0.969(0.019) & 0.895(0.028)- & 0.966(0.013)= & 0.951(0.073)= & 0.959(0.017)= & 0.949(0.012)- & 0.976(0.019)= & 0.923(0.023)- \\
Concrete      & 0.930(0.025) & 0.692(0.040)- & 0.892(0.027)- & 0.883(0.025)- & 0.875(0.025)- & 0.815(0.040)- & 0.939(0.015)= & 0.734(0.056)- \\
Dow chemical   & 0.920(0.020) & 0.696(0.050)- & 0.864(0.016)- & 0.862(0.017)- & 0.855(0.025)- & 0.752(0.055)- & 0.927(0.026)= & 0.785(0.036)- \\
Energy: cooling & 1.111(0.009) & 1.033(0.017)- & 1.092(0.009)- & 1.089(0.012)- & 1.082(0.016)- & 1.052(0.009)- & 1.097(0.016)- & 1.028(0.016)- \\
Energy: heating & 1.144(0.007) & 1.087(0.016)- & 1.128(0.010)- & 1.126(0.009)- & 1.124(0.012)- & 1.101(0.009)- & 1.136(0.009)= & 1.067(0.013)- \\
Tower         & 1.022(0.020)) & 0.812(0.049)- & 0.986(0.035)- & 0.981(0.039)- & 0.985(0.025)- & 0.941(0.037)- & 1.031(0.005)= & 0.859(0.050)- \\
Wine: red       & 0.629(0.059) & 0.591(0.009)- & 0.634(0.009)= & 0.633(0.013)= & 0.632(0.011)= & 0.615(0.014)- & 0.647(0.013)= & 0.613(0.011)- \\
Wine: white     & 0.359(0.074) & 0.354(0.010)= & 0.390(0.020)= & 0.378(0.050)= & 0.387(0.012)= & 0.379(0.006)= & 0.400(0.025)= & 0.361(0.012)= \\
Yacht         & 1.170(0.002) & 1.131(0.024)- & 1.167(0.004)- & 1.167(0.002)- & 1.167(0.003)- & 1.155(0.013)- & 1.172(0.001)+ & 1.137(0.013)- \\
\hline
Total $+/-/=$ & --- &  0/9/1 & 0/7/3 & 0/7/3 & 0/7/3 & 0/9/1 & 1/2/7 & 0/9/1  \\
\bottomrule
\end{tabular}
}
\label{resulttesting}
\end{table*}

\section{Results}\label{section:results}

\subsection{Benchmarking results}

\subsubsection{Results and analysis}
\Cref{resulttraining,resulttesting} show the results obtained when considering the accuracy as measured on the training set and on the test set, respectively, for the parameter settings that result in the average performance.
At training time, evoNSGA-II performs significantly better than any other algorithm in a vast number of cases, sometimes substantially so (e.g., when compared to Adap.~$\alpha$-dom., NSGA-II, and SPEA2 on several data sets). In fact, evoNSGA-II is found to be significantly better than another algorithm 64 times, worse only 1 times, and not significantly different 5 times. When it comes to the test set, evoNSGA-II remains vastly superior, although the number of statistical comparisons that are not significantly different raises to 19 (better 50 times, worse 1 time). This is due to the generalization gap between the training and the test set. In fact, improving generalization is not on focus in this paper.

Overall, only NSGA-II+PD is capable of coming close to the performance of evoNSGA-II.
At training time, evoNSGA-II is better than NSGA-II+PD on 1 data sets, worse on 4, and equal on 5.
At test time, the difference between the two shrinks even more, due to the generalization gap.
Nevertheless, except for in one case (Yacht), evoNSGA-II is essentially equal or better than NSGA-II+PD.

We proceed by briefly describing what we observe for the other parameter configurations.
More detailed information is reported in the Appendix.
Across the algorithms, using a larger population size and larger tournament size contribute to improve the performance, while it is unclear whether using more or less crossover than mutation is preferable.
Across the configurations, evoNSGA-II remains the best-performing approach, although it is sometimes matched by NSGA-II+PD.
However, we observe that with larger population sizes, the gap between evoNSGA-II and NSGA-II+PD grows in favor of the former.
For example, at training time with the best-possible parameter configurations (for both algorithms, using a population size of 5000), evoNSGA-II is significantly superior to NSGA-II+PD on 6 data sets, equal on 4, and worse on none.
This is likely because larger population sizes allow for better estimations of evolvability.

\subsubsection{Further analysis: convergence of HV}
\begin{figure}[]
\centering
\includegraphics[width=\linewidth]{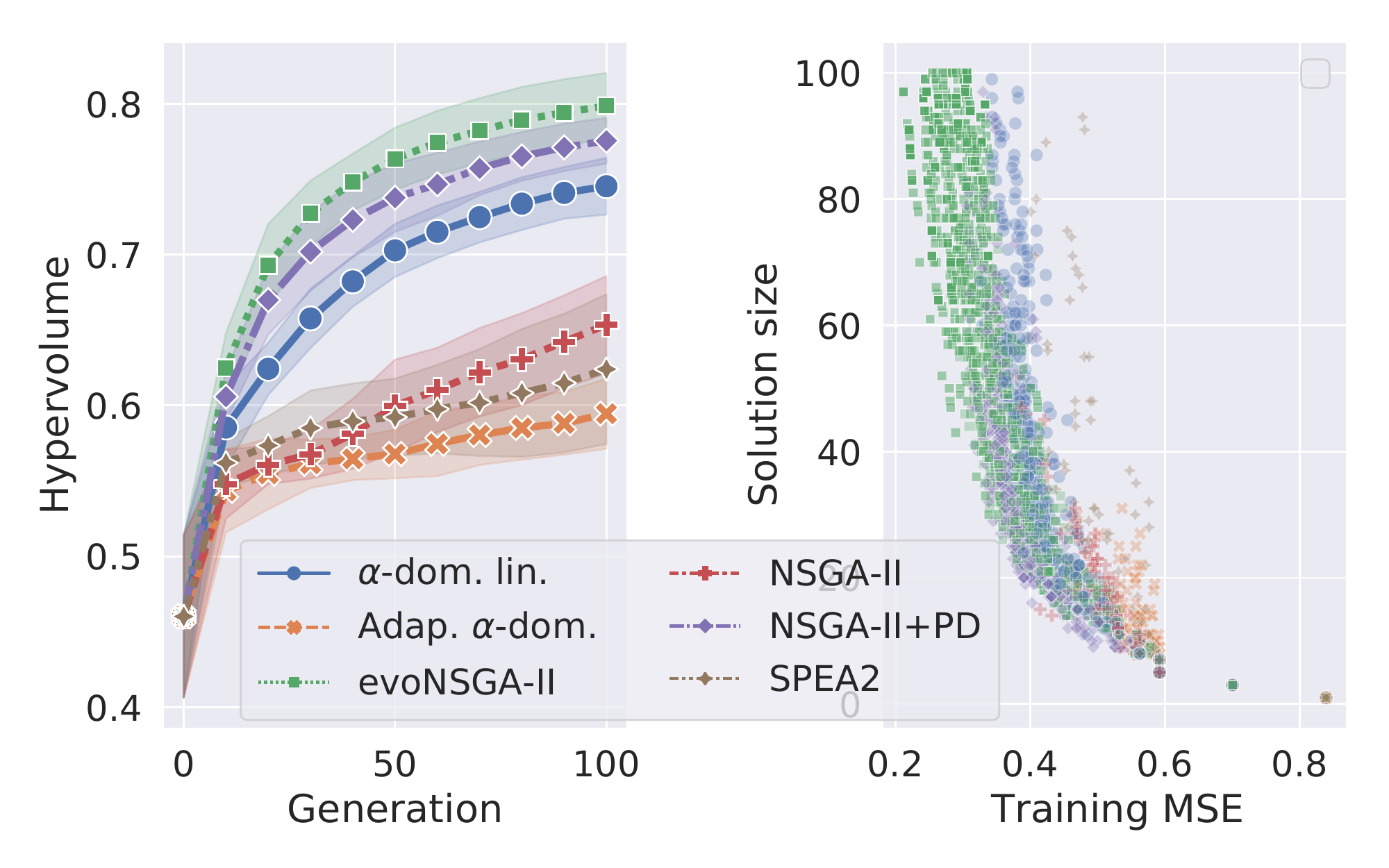}
\caption{Comparison between the algorithms in terms of HV during the evolution (left) and final front (right) for 30 runs on Airfoil at training time.
Left: Lines represent means and shaded areas represent standard deviations.
Right: All solutions in the archives from the 30 runs are shown.
}
\label{HVCurveDuringIteration}
\end{figure}

To provide further evidence that evoNSGA-II is typically superior to the other algorithms, 
\Cref{HVCurveDuringIteration} (left) shows the convergence of the (training) HV, again with using parameter configurations that represent average performance, on Airfoil.
Due to space limitations, we show the respective plots for other data sets in Appendix E.
For clarity, since the non-adaptive $\alpha$-dominance algorithms perform similarly, we report only the one with linear scheduling ($\alpha$-dom.~lin.) in \Cref{HVCurveDuringIteration}.
As can be seen, the HV obtained by NSGA-II, SPEA2, and Adap.~$\alpha$-dom.~tends to converge to a suboptimal value very soon after the first dozen generations. 
The other algorithms, i.e., evoNSGA-II, $\alpha$-dom.~lin., and NSGA-II+PD perform similarly, however evoNSGA-II is slightly superior throughout the whole search.

Furthermore, \Cref{HVCurveDuringIteration} (right) shows the distribution of the solutions  in the final archives (for the 30 runs).
The most apparent result is that only evoNSGA-II is capable of reliably discovering accurate solutions with a larger size than 30 (approximately).
Interestingly, NSGA-II and NSGA-II+PD can be better than evoNSGA-II in discovering some relatively accurate solutions of size between 10 and 20 (approximately).
This is because the search of NSGA-II and NSGA-II+PD can concentrate more in that area, as they discover larger and even more accurate solutions less frequently than evoNSGA-II.

\begin{figure}[!ht]
\centering
\includegraphics[width=0.97\linewidth]{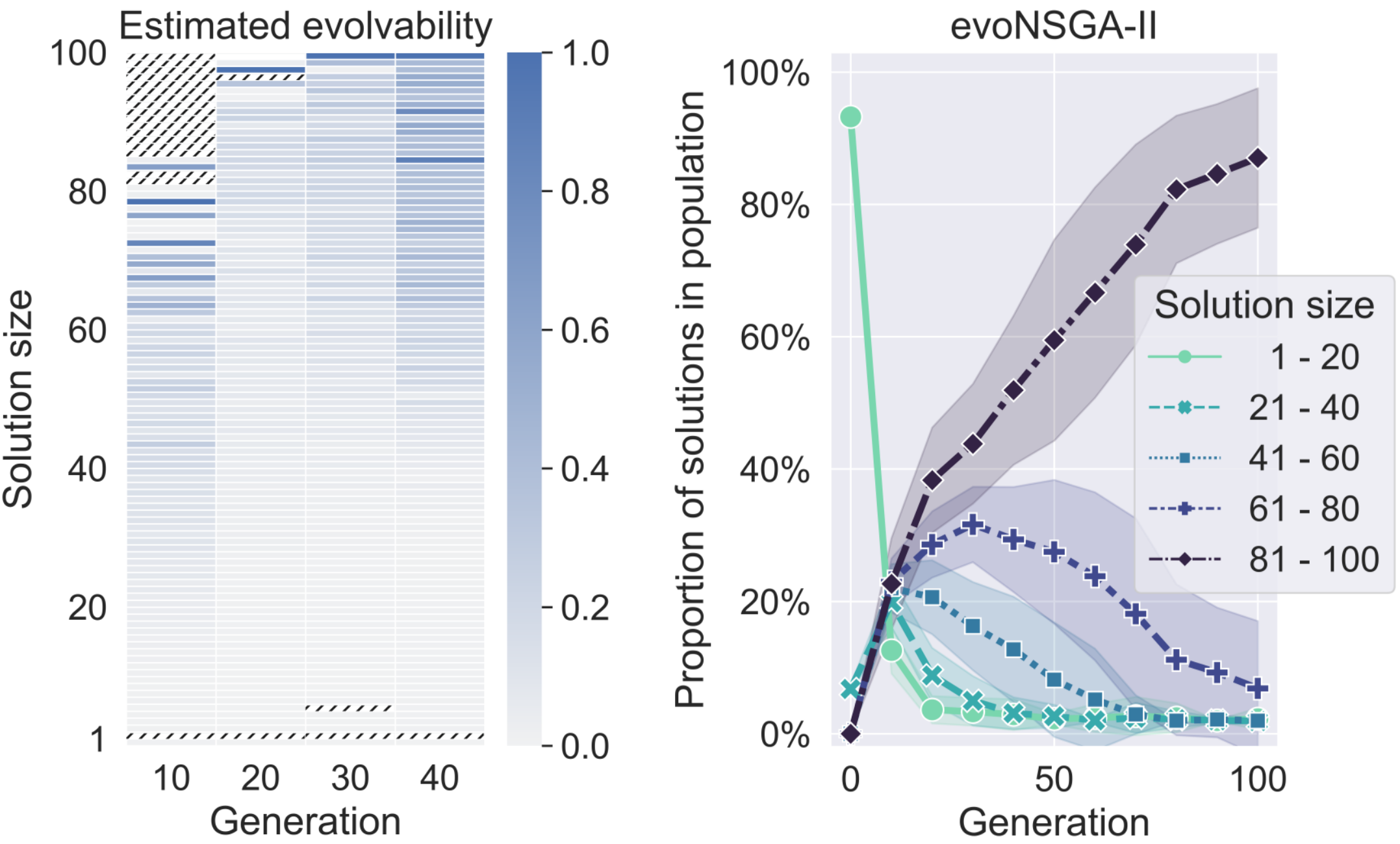}
\caption{Left: Each column of the heat-map shows, for a given generation, the average evolvability between crossover and mutation computed with the workflow of \Cref{sec:evolv-small-solutions} on Airfoil, and normalized across solution size (dashed entries represent absent sizes).
Right: Proportions of solutions of different sizes in evoNSGA-II during 30 evolutions on Airfoil (lines are means, shaded areas are standard deviations).}
\label{MarginalizedHeatMap-explength}
\end{figure}

\subsection{Did it work as expected?}
As last result, we show that evoNSGA-II does not, in fact, exhibit evolvability degeneration.
\Cref{MarginalizedHeatMap-explength} shows, on Airfoil, the evolvability that is estimated using the workflow proposed in \Cref{sec:evolv-small-solutions} (left panel), and also the proportion of solutions of different sizes in the population of evoNSGA-II during the evolution (right panel), again using the parameter configuration that represents average performance.
As can be seen, the proportion of solutions of larger sizes increases over time during the evolution process of evoNSGA-II, which is in agreement with the expected evolvability from our analysis.
This result is in stark contrast with the one displayed in \Cref{fig:NSGAII-forPopulationCollapseAnalysis} (note the different scale in sizes), where NSGA-II could not discover larger and more accurate solutions.

We produced the same plots for the other algorithms and included them in the Appendix D; 
there, it can be seen that SPEA2 and Adap.~$\alpha$~dom., like NSGA-II, suffer from evolvability degeneration.
The other algorithms perform better, yet still assign less copies to larger solutions than evoNSGA-II.



\section{Discussion}\label{section:discussion}

In this work, we investigated \emph{evolvability degeneration}, i.e., the phenomenon by which small solutions over-replicate and hamper search progress because they represent unfruitful parents for the discovery of larger and more accurate offspring.
Next, we proposed to extend NSGA-II into evoNSGA-II, which estimates the evolvability of solutions based on their size and, based on this, bounds how many solutions of any given size can be selected for the next generation.
Lastly, we found that evoNSGA-II is largely superior to other recent MOGP algorithms, and is indeed capable of allowing solutions of highly-evolvable size to thrive.

The reason for evolvability degeneration can be linked to the fact that the algorithm has insufficient time to discover more accurate solutions (because the probability that variation succeeds is low) compared to the speed by which small solutions duplicate.
Such hypothesis is strongly supported by the findings of~\cite{10.1162/evco.1998.6.4.293}, where it is shown that GP tends to fail when the pressure surpasses a certain threshold.
It is thus natural that MOGP algorithms that improve the diversity of the population perform better than classic NSGA-II.
In fact, the algorithms that we used in our comparisons that were built to improve NSGA-II, essentially realize some form of diversity preservation.
However, none of them considers tracking and using evolvability to decide which solutions to keep and which to discard.
In our view, this is the fundamental reason why evoNSGA-II performed best.
Interestingly, NSGA-II+PD, which is perhaps an even simpler approach than evoNSGA-II, performed similarly to evoNSGA-II in many data sets.
Still, evoNSGA-II performed typically equal to or better than NSGA-II+PD, suggesting that one may not always want to discard all duplicate solutions: keeping a number of copies for highly-evolvable solutions seems to be generally more helfpul.
Moreover, we observed that the performance gap between evoNSGA-II and NSGA-II+PD tends to increase when the population size is larger.
We believe that this happens because larger population sizes allow for better estimations of evolvability.

There exist a number of limitations in this paper that call for future research.
Firstly, our estimations of evolvability are repeated every generation, using solely the current population.
We attempted to use exponential-moving-averages to incorporate estimations from previous generations but preliminary findings indicated no statistically significant improvement.
However, one could study whether other approaches can lead to an improvement, such as learning an accurate model of evolvability of solution size across multiple data sets and parameter configurations, and using that model as starting point when dealing with a new problem.
A second important limitation is that we considered minimizing solution size, which is a simple but coarse way of pursuing interpretability. 
Future work should consider other and better proxies of interpretability (e.g.,~\cite{burlacu2019parsimony,Virgolin2020PPSN}), and assess whether the good performance found here for evoNSGA-II transfers to those settings.
Transferability of the quality of our approach should also be assessed when other variation operators are used, such as geometric semantic-~\cite{moraglio2012geometric,pawlak2014semantic} or linkage-based ones~\cite{virgolin2021improving}, as well as, e.g., gradient descent ot optimize coefficients~\cite{dick2020feature}.
A third limitation is that evoNSGA-II makes no attempt to limit bloat. 
If bloated solutions have larger evolvability, they will replicate more than others.
For SR this is not necessarily a problem, since it is reasonable to impose a cap on the maximally allowed size, above which solutions would certaintly not be interpretable.
However, capping the size might not be desirable for other problems.
There, evoNSGA-II might keep discovering larger and larger solutions, and thus fail to find medium-sized ones.
Thus, bloat-control mechanisms may need to be considered.

Finally, we conclude this work by reflecting on the fact that our results may, in principle, transfer to problems of very different nature than GP for SR.
Indeed, we remark that maximizing accuracy and minimizing solution size is an \emph{imbalanced} multi-objective problem:
on the one hand, minimizing solution size is easily done, since random deletion of components suffices to improve this objective; 
on the other hand, maximizing accuracy is almost always challenging, since the right components need to appear in the right order to obtain an accurate model. 
There might exist a number of problems where a similar situation happens, i.e., the objective that is easy-to-optimize inhibits the search of solutions with respect to the objective that is hard-to-optimize.
For any given problem, tracking and exploiting information on the evolvability in terms of the hard-to-optimize objective that is associated with the easy-to-optimize objective might be a consistent way to improve multi-objective evolutionary search.

\section{Conclusion}\label{sec:conclusion}
We studied an important cause of inefficiency in the use of the non-dominated sorting genetic algorithm II (NSGA-II) for the discovery of symbolic regression models with trade-offs between accuracy and simplicity.
Namely, we experimentally found that simpler models over-replicate and take over the majority of the population, because they lack \emph{evolvability}, i.e., they represent infertile grounds for larger but more accurate models to be discovered.
We named this phenomenon \emph{evolvability degeneration}, and proposed \emph{evoNSGA-II}, an algorithm that is explicitly built to prevent it.
With comparisons to NSGA-II and six other algorithms, upon ten real-world data sets, and across different parameter configurations, we found evoNSGA-II to be the superior approach.
The working principles of evoNSGA-II are not limited to symbolic regression: studying their transferability to other imbalanced multi-objective problems represents an interesting avenue for future research.

\begin{acks}
This research was funded by the European Commission within the HORIZON Programme (Trust AI Project, Contract No.:952060). We further thank Maurits and Anna de Kock Foundation for financing a high-performance computing system.
\end{acks}


\appendix

\section*{Appendix}

\section{Survey on the popularity of NSGA-II}

We performed a small survey to assess the popularity of NSGA-II as a recent approach for multi-objective genetic programming (MOGP).
The survey was carried out considering the following journals and conferences:
\begin{itemize}
\item Evolutionary Computation Journal;
\item European Conference on Genetic Programming (EuroGP);
\item Genetic Programming and Evolvable Machines;
\item IEEE Congress on Evolutionary Computatin Evolvable Machines (IEEE CEC);
\item IEEE Transactions on Cybernetics;
\item IEEE Transactions on Evolutionary Computation;
\item International Conference on Evolutionary Multi-Criterion Optimization (EMO);
\item International Conference on Parallel Problem Solving from Nature (PPSN);
\item Swarm and Evolutionary Computation;
\item The Genetic and Evolutionary Computation Conference (GECCO).
\end{itemize}

We searched for works on NSGA-II for MOGP on Scopus (\url{www.scopus.com}) provided by Elsevier, using the query listed in \Cref{SCOPUSSTRING}.

\begin{table*}[h]
\caption{Query used in SCOPUS to compile the short survey on the use of NSGA-II in MOGP. Keywords ``VENUE1'', ``VENUE2'', etc.~represent the names of journals (for which the field SOURCE-TITLE is used for the search) and conferences (for which the field CONFERENCE is used for the search) from the list provided above.}
\label{SCOPUSSTRING}
\small
\begin{tabular}{cc}
\toprule
VenueQuery & \small  SOURCE-TITLE(VENUE1) OR CONFERENCE(VENUE2) OR \dots \\
\cline{2-2} 
\multirow{3}{*}{MultiObjQuery} & KEY ( ( multi-objective ) OR ( multi AND objective ) \\
 & OR ( multiobjective )OR ( many-objective ) \\
 & OR ( many AND objective ) OR ( manyobjective ) )\\
\cline{2-2} 
 \multirow{2}{*}{GPorSRQuery} & KEY ( ( genetic AND programming ) \\
 & OR ( symbolic AND regression ) ) )\\
 \cline{2-2} 
 YearQuery & PUBYEAR > 2016 AND PUBYEAR < 2022\\
\midrule
\multirow{2}{*}{Filter criteria} & VenueQuery AND MultiObjQuery AND \\
 & GPorSRQuery AND YearQuery\\
\bottomrule
\end{tabular}
\end{table*}

Ultimately, we obtained \Cref{SurveyOfMOGP}.
The column \#MOGP indicates the number of works matched by the query in \Cref{SCOPUSSTRING}.
The column \#NSGA-II indicates the number of works that \emph{use} NSGA-II as MOGP method.
We consider a work to \emph{use} NSGA-II if NSGA-II is effectively run (either as the main approach, or as a baseline approach). 
If NSGA-II is cited but not run, then we do not consider it to be used.
Overall, NSGA-II is a very popular MOGP method, as it is used on approximately $70\%$ of the considered works.
For completeness, all works are listed in the references at the end of this document~\cite{thite2021concurrent,zille2021unit,raymond2021multi,ohki2020benchmark,wang2021two,saidani2020prediction,bi2020automatically,weiner2020automatic,kubalik2020symbolic,galvan2019promoting,roman2019sentiment,fraser2019emocs,yuan2019hybrid,evans2019s,vu2019toward,burlacu2019parsimony,la2018analysis,escobar2018multi,watanabe2018analyzing,wu2018evolutionary,gomez2017hyper,da2017fragment,sadowski2017exploring,yuan2021expensive,nag2019feature,gong2019similarity,li2020multi,lai2020solving,saad2018multi,meghwani2017multi,bokhari2020genetic,yang2020multi,golabi2020bypassing,Silva2020APM,aldeia2020parametric,masood2020fitness,ojha2020multi,grochol2020evolutionary,wang2020multi,souza2020novel,zhang2019evolving,seyam2019multi,sun2019preference,burlacu2019online,xue2018genetic,nyman2018metaheuristics,tahmassebi2018pareto,hammami2018multi,oliveira2018multi,fracasso2018multi,wan2021multiobjective,wang2018decomposition,li2016multiobjective,li2015bi,la2019probabilistic,mariot2021evolutionary,rockett2021constant,lensen2020multi,santana2019gp,kim2019software,dou2018comparison,10.1007/s10710-017-9310-3,grochol2018multi,minarik2017evolutionary,pinos2021evolutionary,vu2021using,azimlu2019designing}.

\begin{table*}[h]
\caption{Result of our brief survey on the popularity of NSGA-II as MOGP algorithm in recent years.
The column \#MOGP reports the number of papers we found that are about MOGP, \#NSGA-II how many of those papers use NSGA-II, and Ratio reports the latter over the former.}
\label{SurveyOfMOGP}
\begin{tabular}{cccc}
\toprule
Year & \#MOGP & \#NSGA-II & Ratio \\
\midrule
2021 & 10 & 7 & 70.0\% \\
2020 & 20 & 13 & 65.0\% \\
2019 & 13 & 9 & 69.2\% \\
2018 & 18 & 13 & 72.2\% \\
2017 & 6 & 3 & 50.0\% \\
\midrule
Total & 67 & 45 & 67.2\% \\
\bottomrule
\end{tabular}
\end{table*}

\section{Information on the datasets}
The number of samples and attributes about datasets used in the comparison study are summarized in \Cref{tab:summarize-of-dataset-information}. 


\begin{table*}[h]
\caption{Summary of the data sets considered in this work. For some data sets, we could not find the original link: those links will be updated to point to our own repository after double-blind review.}
\label{tab:summarize-of-dataset-information}
\begin{tabular}{lccl}
\toprule
Name & \#Samples & \#Attributes & Link\\
\midrule
Airfoil & 1503 & 6 & \Small \url{https://archive.ics.uci.edu/ml/datasets/airfoil+self-noise}\\
Boston  & 506 & 14 & \Small \url{https://archive.ics.uci.edu/ml/machine-learning-databases/housing}\\
Concrete  & 1030 & 9 & \Small \url{https://archive.ics.uci.edu/ml/datasets/concrete+compressive+strength}\\
Dow chemical & 1066 & 58 & \Small \emph{original-link-missing}\\
Energy: cooling & 768 & 8 & \Small
\url{https://archive.ics.uci.edu/ml/datasets/energy+efficiency}\\
Energy: heating & 768 & 8 & \Small
\url{https://archive.ics.uci.edu/ml/datasets/energy+efficiency}\\
Tower & 4999 & 26 & \Small
\emph{original-link-missing}\\
Wine: red & 1599 & 12 & \Small
\url{https://archive.ics.uci.edu/ml/datasets/wine+quality}\\
Wine: white & 4898 & 12 & \Small
\url{https://archive.ics.uci.edu/ml/datasets/wine+quality}\\
Yacht & 308 & 7 & \Small
\url{https://archive.ics.uci.edu/ml/datasets/yacht+hydrodynamics}\\
\bottomrule
\end{tabular}
\end{table*}

\section{Hyper-volume results under best and worse parameter settings}

We begin by reporting, in \Cref{tab:best-median-worse-parameter-settings}, the parameter configurations that result, for each algorithm, in worst, average, and best performance with respect to training HVs.
The way in which we compute how parameter configurations fare with respect to each other is explained in Sec.~5 of the paper.
Note that due to the high computational cost of SPEA2, we did not run the algorithm for a population size of 5000. 
Hence, the parameter configuration that represents average performance cannot be computed. 
We therefore decided to set it to a population size of 1000, since this population size is the most frequent for the configurations that represent the average performance for the other algorithms, and the crossover-mutation proportion that leads to the best performance under that population size (tournament selection is not used in SPEA2).
Since we do not have results for a population size of 5000 for SPEA2, we do not consider this algorithm in terms of best-case performance.

\begin{table*}[h]
\caption{Parameter settings of all comparison methods that lead to worst, average and best performance. For SPEA2, tournament size does not apply.}
\resizebox{\textwidth}{!}{
\begin{tabular}{l ccc ccc ccc}
\toprule
 &
  \multicolumn{3}{c}{Worst case} &
  \multicolumn{3}{c}{Average case} &
  \multicolumn{3}{c}{Best case} \\
Algorithm &
  \multicolumn{1}{c}{Pop.~size} &
  \multicolumn{1}{c}{Cross.-Mut.} &
  Tourn.~size &
  \multicolumn{1}{c}{Pop.~size} &
  \multicolumn{1}{c}{Cross.-Mut.} &
  Tourn.~size &
  \multicolumn{1}{c}{Pop.~size} &
  \multicolumn{1}{c}{Cross.-Mut.} &
  Tourn.~size \\ \hline
evoNSGA-II &
  \multicolumn{1}{c}{250} &
  \multicolumn{1}{c}{0.5-0.5} &
  1 &
  \multicolumn{1}{c}{1000} &
  \multicolumn{1}{c}{0.9-0.1} &
  2 &
  \multicolumn{1}{c}{5000} &
  \multicolumn{1}{c}{0.9-0.1} &
  7 \\
Adap.~$\alpha$-dom. &
  \multicolumn{1}{c}{250} &
  \multicolumn{1}{c}{0.5-0.5} &
  1 &
  \multicolumn{1}{c}{1000} &
  \multicolumn{1}{c}{0.9-0.1} &
  1 &
  \multicolumn{1}{c}{5000} &
  \multicolumn{1}{c}{0.9-0.1} &
  7 \\
$\alpha$-dom.~cos. &
  \multicolumn{1}{c}{250} &
  \multicolumn{1}{c}{0.5-0.5} &
  1 &
  \multicolumn{1}{c}{250} &
  \multicolumn{1}{c}{0.9-0.1} &
  7 &
  \multicolumn{1}{c}{5000} &
  \multicolumn{1}{c}{0.5-0.5} &
  7 \\
$\alpha$-dom.~lin. &
  \multicolumn{1}{c}{250} &
  \multicolumn{1}{c}{0.5-0.5} &
  1 &
  \multicolumn{1}{c}{1000} &
  \multicolumn{1}{c}{0.5-0.5} &
  7 &
  \multicolumn{1}{c}{5000} &
  \multicolumn{1}{c}{0.5-0.5} &
  7 \\
$\alpha$-dom.~sigm. &
  \multicolumn{1}{c}{250} &
  \multicolumn{1}{c}{0.5-0.5} &
  1 &
  \multicolumn{1}{c}{1000} &
  \multicolumn{1}{c}{0.9-0.1} &
  2 &
  \multicolumn{1}{c}{5000} &
  \multicolumn{1}{c}{0.5-0.5} &
  7 \\
NSGA-II &
  \multicolumn{1}{c}{250} &
  \multicolumn{1}{c}{0.5-0.5} &
  1 &
  \multicolumn{1}{c}{1000} &
  \multicolumn{1}{c}{0.9-0.1} &
  2 &
  \multicolumn{1}{c}{5000} &
  \multicolumn{1}{c}{0.5-0.5} &
  7 \\
NSGA-II+PD &
  \multicolumn{1}{c}{250} &
  \multicolumn{1}{c}{0.9-0.1} &
  1 &
  \multicolumn{1}{c}{1000} &
  \multicolumn{1}{c}{0.9-0.1} &
  2 &
  \multicolumn{1}{c}{5000} &
  \multicolumn{1}{c}{0.5-0.5} &
  7 \\
SPEA2 &
  \multicolumn{1}{c}{250} &
  \multicolumn{1}{c}{0.9-0.1} &
  - &
  \multicolumn{1}{c}{1000} &
  \multicolumn{1}{c}{0.5-0.5} &
  - &
  - &
  - &
  - \\ \hline
\end{tabular}
}
\label{tab:best-median-worse-parameter-settings}

\end{table*}

In the paper, we show results for the parameter configurations that represent average performance.
Here we additional show the results obtained for the configurations that represent worst and best performance. 
The training and test results for worst-case are respectively shown in \Cref{tab:resulttraining-250-100-0.5-0.5-1,tab:resulttesting-250-100-0.5-0.5-1}.
The training and test results for best-case are respectively shown in \Cref{tab:resulttraining-5000-100-0.9-0.1-7,tab:resulttesting-5000-100-0.9-0.1-7}.

The most important result is that evoNSGA-II is still typically preferable to the other algorithms at training time.
As said in the paper, we consider training results to be more relevant because we made no attempt to improve generalization performance here, rather, our goal was to improve search efficiency (by contrasting evolvability degeneration).
With the configurations that represent worst performance, which always use the smallest population size of 250 (cfr.~\Cref{tab:best-median-worse-parameter-settings}), evoNSGA-II and NSGA-II+PD are essentially equivalent. 
With the configurations that represt best performance, which always use the largest population size of 5000, evoNSGA-II outperforms also NSGA-II+PD on more datasets than for the other configurations. 
We believe this is a consequence of the fact that, with the largest population size, the estimations of evolvability are more accurate. 
At test time, evoNSGA-II remains preferable in many cases, except when compared with NSGA-II+PD, which performs better than evoNSGA-II for worst-case scenarios, or similarly for best-case scenarios.
Since evoNSGA-II obtains a better training performance than NSGA-II+PD, we can expect that including mechanisms to improve generalization will benefit evoNSGA-II more than NSGA-II+PD.
As mentioned above, the result of SPEA2 with population size 5000 is omitted due to the high computational cost of running this algorithm.

\begin{table*}[h]
\caption{Mean (standard deviation) of the HV computed on the training set for 30 runs of the considered algorithms.
This table corresponds to the parameter configurations that represent worst performance.
The symbols $+,-,=$ indicate, for each algorithm other than evoNSGA-II, whether the corresponding distribution of results for evoNSGA-II is, respectively, significantly better, worse, or not significantly different.
The last row summarizes this information.
}
\resizebox{\textwidth}{!}{
\begin{tabular}{lcccccccc}
\toprule
Dataset & evoNSGA-II & Adap. $\alpha$-dom. & $\alpha$-dom. cos. & $\alpha$-dom. lin. & $\alpha$-dom. sig. & NSGA-II & NSGA-II+PD & SPEA2 \\
\midrule
Airfoil       & 0.731(0.026) & 0.564(0.019)- & 0.564(0.019)- & 0.564(0.019)- & 0.564(0.019)- & 0.589(0.043)- & 0.720(0.017)= & 0.596(0.040)- \\
Boston        & 0.983(0.016) & 0.866(0.024)- & 0.866(0.024)- & 0.866(0.024)- & 0.866(0.024)- & 0.923(0.035)- & 0.987(0.011)= & 0.888(0.021)- \\
Concrete      & 0.877(0.034) & 0.636(0.031)- & 0.636(0.031)- & 0.636(0.031)- & 0.636(0.031)- & 0.711(0.064)- & 0.890(0.026)= & 0.669(0.050)- \\
Dow chemical   & 0.904(0.023) & 0.664(0.042)- & 0.664(0.042)- & 0.664(0.042)- & 0.664(0.042)- & 0.744(0.076)- & 0.934(0.014)+ & 0.771(0.062)- \\
Energy: cooling & 1.108(0.015) & 1.032(0.015)- & 1.032(0.015)- & 1.032(0.015)- & 1.032(0.015)- & 1.043(0.015)- & 1.075(0.014)- & 1.020(0.008)- \\
Energy: heating & 1.134(0.013) & 1.050(0.020)- & 1.050(0.020)- & 1.050(0.020)- & 1.050(0.020)- & 1.079(0.023)- & 1.111(0.013)- & 1.029(0.017)- \\
Tower         & 0.962(0.033) & 0.767(0.031)- & 0.767(0.031)- & 0.767(0.031)- & 0.767(0.031)- & 0.852(0.059)- & 1.003(0.009)+ & 0.813(0.048)- \\
Wine: red       & 0.493(0.012) & 0.440(0.002)- & 0.440(0.002)- & 0.440(0.002)- & 0.440(0.002)- & 0.449(0.013)- & 0.493(0.006)= & 0.451(0.014)- \\
Wine: white     & 0.422(0.016) & 0.361(0.014)- & 0.361(0.014)- & 0.361(0.014)- & 0.361(0.014)- & 0.372(0.021)- & 0.430(0.007)= & 0.371(0.007)- \\
Yacht         & 1.176(0.010) & 1.112(0.044)- & 1.112(0.044)- & 1.112(0.044)- & 1.112(0.044)- & 1.126(0.052)- & 1.173(0.002)= & 1.122(0.033)- \\
\hline
Total $+/-/=$ & --- &  0/10/0 & 0/10/0 & 0/10/0 & 0/10/0 & 0/10/0 & 2/2/6 & 0/10/0 \\
\bottomrule
\end{tabular}
}
\label{tab:resulttraining-250-100-0.5-0.5-1}
\end{table*}

\begin{table*}[h]
\caption{Results for the test set, formatting similar to that of \Cref{tab:resulttraining-250-100-0.5-0.5-1}.
}
\resizebox{\textwidth}{!}{
\begin{tabular}{lcccccccc}
\toprule
Dataset & evoNSGA-II & Adap. $\alpha$-dom. & $\alpha$-dom. cos. & $\alpha$-dom. lin. & $\alpha$-dom. sig. & NSGA-II & NSGA-II+PD & SPEA2 \\
\midrule
Airfoil       & 0.761(0.027) & 0.634(0.018)- & 0.634(0.018)- & 0.634(0.018)- & 0.634(0.018)- & 0.658(0.042)- & 0.763(0.021)= & 0.671(0.038)- \\
Boston        & 0.951(0.022) & 0.868(0.024)- & 0.868(0.024)- & 0.868(0.024)- & 0.868(0.024)- & 0.925(0.030)- & 0.973(0.013)+ & 0.888(0.026)- \\
Concrete      & 0.872(0.042) & 0.643(0.031)- & 0.643(0.031)- & 0.643(0.031)- & 0.643(0.031)- & 0.729(0.070)- & 0.900(0.019)= & 0.683(0.056)- \\
Dow chemical   & 0.840(0.041) & 0.619(0.066)- & 0.619(0.066)- & 0.619(0.066)- & 0.619(0.066)- & 0.719(0.070)- & 0.883(0.019)+ & 0.732(0.059)- \\
Energy: cooling & 1.088(0.015) & 1.014(0.019)- & 1.014(0.019)- & 1.014(0.019)- & 1.014(0.019)- & 1.035(0.018)- & 1.072(0.015)- & 1.003(0.011)- \\
Energy: heating & 1.122(0.013) & 1.058(0.019)- & 1.058(0.019)- & 1.058(0.019)- & 1.058(0.019)- & 1.083(0.022)- & 1.120(0.011)= & 1.042(0.016)- \\
Tower         & 0.957(0.035) & 0.754(0.030)- & 0.754(0.030)- & 0.754(0.030)- & 0.754(0.030)- & 0.846(0.062)- & 1.004(0.010)+ & 0.806(0.049)- \\
Wine: red       & 0.623(0.021) & 0.587(0.004)- & 0.587(0.004)- & 0.587(0.004)- & 0.587(0.004)- & 0.596(0.016)- & 0.638(0.010)= & 0.600(0.017)- \\
Wine: white     & 0.362(0.063) & 0.330(0.023)- & 0.330(0.023)- & 0.330(0.023)- & 0.330(0.023)- & 0.350(0.025)- & 0.392(0.018)= & 0.350(0.014)- \\
Yacht         & 1.155(0.019) & 1.095(0.055)- & 1.095(0.055)- & 1.095(0.055)- & 1.095(0.055)- & 1.107(0.067)- & 1.166(0.003)+ & 1.101(0.042)- \\
\hline
Total $+/-/=$ & --- & 0/10/0 & 0/10/0 & 0/10/0 & 0/10/0 & 0/10/0 &  4/1/5 & 0/10/0\\
\bottomrule
\end{tabular}
}
\label{tab:resulttesting-250-100-0.5-0.5-1}
\end{table*}

\begin{table*}
\caption{Mean (standard deviation) of the HV computed on the training set for 30 runs of the considered algorithms.
This table corresponds to the parameter configurations that represent best performance.
The symbols $+,-,=$ indicate, for each algorithm other than evoNSGA-II, whether the corresponding distribution of results for evoNSGA-II is, respectively, significantly better, worse, or not significantly different.
The last row summarizes this information.
}
\resizebox{\textwidth}{!}{
\begin{tabular}{lcccccccc}
\toprule
  Dataset & evoNSGA-II & Adap. $\alpha$-dom. & $\alpha$-dom. cos. & $\alpha$-dom. lin. & $\alpha$-dom. sig. & NSGA-II & NSGA-II+PD & SPEA2 \\
\midrule
Airfoil       & 0.843(0.015) & 0.629(0.032)- & 0.821(0.017)- & 0.816(0.014)- & 0.822(0.016)- & 0.727(0.026)- & 0.818(0.017)- & - \\
Boston        & 1.057(0.007) & 0.926(0.021)- & 1.037(0.008)- & 1.036(0.009)- & 1.042(0.008)- & 0.995(0.010)- & 1.050(0.007)- & - \\
Concrete      & 0.969(0.011) & 0.751(0.059)- & 0.950(0.011)- & 0.943(0.015)- & 0.952(0.014)- & 0.897(0.029)- & 0.967(0.016)= & - \\
Dow chemical   & 1.008(0.007) & 0.773(0.065)- & 0.989(0.007)- & 0.983(0.006)- & 0.994(0.005)- & 0.872(0.036)- & 1.011(0.005)= & - \\
Energy: cooling & 1.132(0.003) & 1.064(0.015)- & 1.127(0.004)- & 1.125(0.005)- & 1.129(0.005)= & 1.080(0.016)- & 1.123(0.009)- & - \\
Energy: heating & 1.164(0.003) & 1.093(0.011)- & 1.150(0.004)- & 1.150(0.004)- & 1.154(0.004)- & 1.119(0.016)- & 1.151(0.005)- & - \\
Tower         & 1.058(0.006)) & 0.923(0.046)- & 1.041(0.007)- & 1.038(0.007)- & 1.044(0.005)- & 0.992(0.022)- & 1.058(0.006)= & - \\
Wine: red       & 0.531(0.005) & 0.459(0.012)- & 0.521(0.004)- & 0.522(0.005)- & 0.523(0.004)- & 0.485(0.010)- & 0.525(0.004)- & - \\
Wine: white     & 0.467(0.003) & 0.386(0.014)- & 0.455(0.004)- & 0.454(0.003)- & 0.457(0.003)- & 0.419(0.010)- & 0.460(0.002)- & - \\
Yacht         & 1.178(0.002) & 1.162(0.011)- & 1.178(0.001)- & 1.177(0.001)- & 1.178(0.001)= & 1.170(0.007)- & 1.179(0.000)= & - \\
\hline
Total $+/-/=$ & --- & 0/10/0 & 0/10/0 & 0/10/0 & 0/8/2 & 0/10/0 & 0/6/4 & - \\
\bottomrule
\end{tabular}
}
\label{tab:resulttraining-5000-100-0.9-0.1-7}
\end{table*}

\begin{table*}[h]
\caption{Results for the test set, formatting similar to that of \Cref{tab:resulttraining-5000-100-0.9-0.1-7}.
}
\resizebox{\textwidth}{!}{
\begin{tabular}{lcccccccc}
\toprule
  Dataset & evoNSGA-II & Adap. $\alpha$-dom. & $\alpha$-dom. cos. & $\alpha$-dom. lin. & $\alpha$-dom. sig. & NSGA-II & NSGA-II+PD & SPEA2 \\
\midrule
Airfoil       & 0.836(0.029) & 0.696(0.026)- & 0.825(0.017)= & 0.825(0.018)= & 0.828(0.020)= & 0.768(0.029)- & 0.821(0.021)= & - \\
Boston        & 0.919(0.149) & 0.922(0.039)+ & 0.918(0.168)= & 0.944(0.106)= & 0.974(0.019)= & 0.964(0.016)= & 0.964(0.092)= & - \\
Concrete      & 0.959(0.010)) & 0.760(0.061)- & 0.940(0.015)- & 0.936(0.014)- & 0.944(0.013)- & 0.905(0.023)- & 0.957(0.011)= & - \\
Dow chemical   & 0.962(0.013) & 0.745(0.056)- & 0.940(0.020)- & 0.928(0.027)- & 0.940(0.022)- & 0.827(0.032)- & 0.966(0.015)= & - \\
Energy: cooling & 1.122(0.005) & 1.056(0.018)- & 1.118(0.005)= & 1.118(0.006)= & 1.121(0.006)= & 1.075(0.018)- & 1.119(0.007)= & - \\
Energy: heating & 1.158(0.007) & 1.104(0.010)- & 1.152(0.004)- & 1.152(0.005)- & 1.155(0.003)= & 1.120(0.014)- & 1.154(0.006)= & - \\
Tower         & 1.053(0.009) & 0.918(0.053)- & 1.015(0.096)- & 1.014(0.065)- & 1.033(0.030)- & 0.994(0.028)- & 1.057(0.005)= & - \\
Wine: red       & 0.641(0.012) & 0.610(0.017)- & 0.634(0.014)= & 0.620(0.070)= & 0.638(0.019)= & 0.640(0.011)= & 0.647(0.008)= & - \\
Wine: white     & 0.360(0.087) & 0.365(0.012)= & 0.303(0.091)= & 0.308(0.085)= & 0.334(0.067)= & 0.390(0.009)= & 0.373(0.060)= & - \\
Yacht         & 1.173(0.003) & 1.155(0.010)- & 1.172(0.002)= & 1.172(0.002)= & 1.173(0.002)= & 1.162(0.009)- & 1.175(0.001)= & - \\
\hline
Total $+/-/=$ & --- &  1/8/1 & 0/4/6 & 0/4/6 & 0/3/7 & 0/7/3 & 0/0/10 & - \\
\bottomrule
\end{tabular}
}
\label{tab:resulttesting-5000-100-0.9-0.1-7}
\end{table*}

\section{Proportion of solution sizes during the evolution for all comparison algorithms}

\begin{figure*}[ht]
    \centering
    \begin{tabular}{c}
        \hspace{1cm} \includegraphics[width=0.91\linewidth]{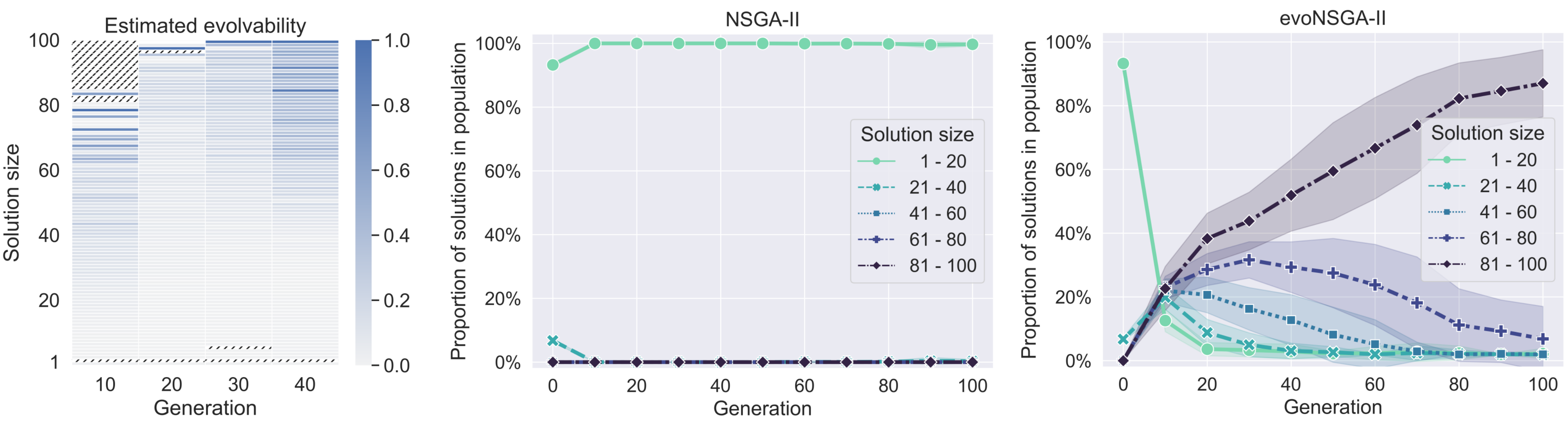}\\
        \includegraphics[width=\linewidth]{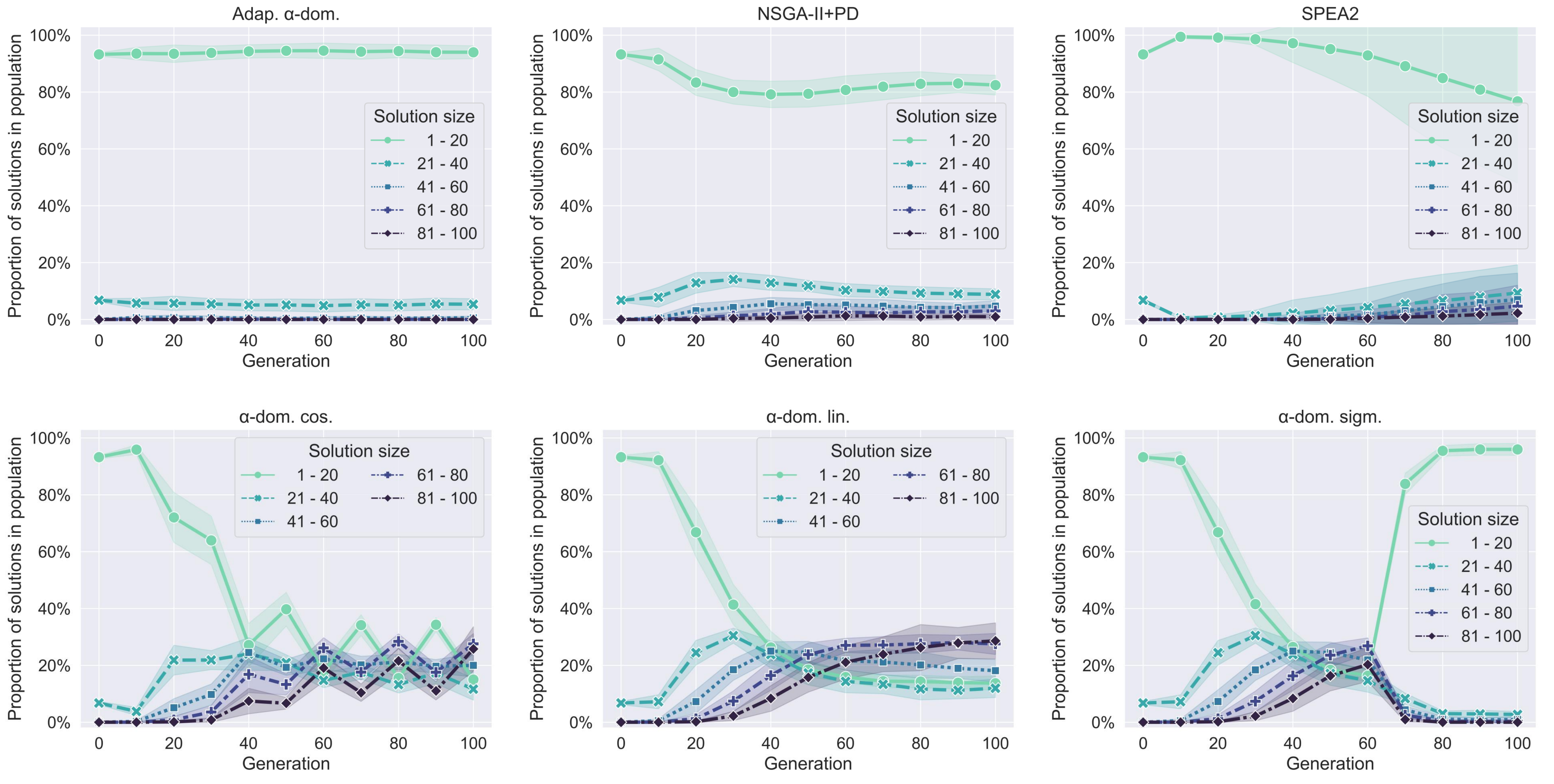}\\ 
    \end{tabular}
    \caption{
    \textbf{Top-left}: Each column of the heat-map shows, for a given generation, the average evolvability between crossover and mutation computed with the workflow described in Section 2 of the paper for the data set Airfoil.
    Values in each column are normalized (dashed entries represent absent values). \textbf{Other plots}: Proportion of solutions of different sizes for the considered algorithms during the evolution on Airfoil. }
    \label{fig:expression-length-and-heatmap-for-supplymaterial}
\end{figure*}

In this section we present the proportion of solution sizes that co-exist in the population during the evolution on the Airfoil data set for all considered algorithms, under the parameter configurations that represent average performance. 
The heat-map of estimated evolvability (computed with the workflow described in Sec.~2 of the paper) is also presented here.
We report the heat-map in terms of average estimated evolvability between crossover and mutation, when computed at different generation limits, namely 10, 20, 30, and 40.
These results are displayed in \Cref{fig:expression-length-and-heatmap-for-supplymaterial}

Among all eight comparison algorithms, evoNSGA-II tends to search for small solutions at the initial stage and then gives more importance to large solutions. 
For the other algorithms, small solutions often occupy a large extent of the population through the entire evolution. 
When it comes to the (non-adaptive) $\alpha$-dominance methods, the growth of large solutions is dictated by the chosen schedule.
Note that small solutions are relatively abundant at initialization anyway, due to the use of the ramped half-and-half scheme.

\section{Hyper-volume on the training set during the evolution}

\begin{figure*}[h]
\centering
\includegraphics[width=\linewidth]{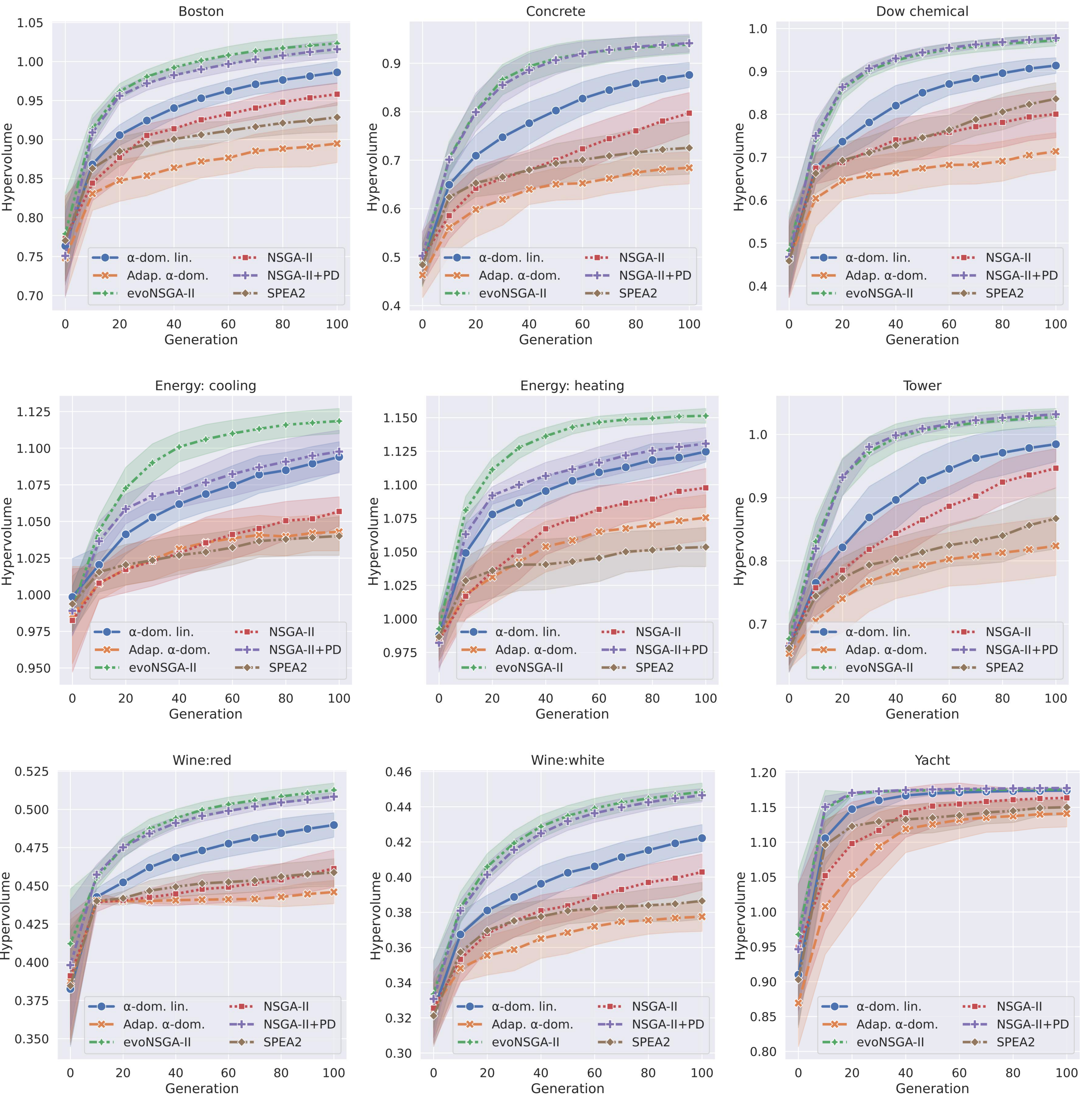}
\caption{HV on the training set for 30 runs on all the considered datasets under average-case parameter settings for all algorithms (means are lines, standard deviations are shaded areas).}
\label{fig:HV-curves-for-all-datasets}
\end{figure*}

\Cref{fig:HV-curves-for-all-datasets} shows the training hyper-volume (HV) during the evolution using the parameter configurations that represent average performance across all data sets, except for Airfoil (the respective plot is already reported in the paper).
It can be seen that evoNSGA-II is almost always the best method consistently through the entirety of the evolution.
NSGA-II+PD is the main contender to evoNSGA-II.
For example, NSGA-II matches the performance of evoNSGA-II on Concrete, Dow chemical, and Yacht. 
However, evoNSGA-II performs better than NSGA-II+PD on the other data sets, sometimes notably so, as on the two Energy data sets.

\newpage 
\bibliography{arxiv}

\end{document}